\documentclass[letterpaper]{article}
\usepackage{aaai2027}
\usepackage[hyphens]{url}
\usepackage{graphicx}
\usepackage{natbib}
\usepackage{caption}
\usepackage{amsmath}
\usepackage{amssymb}
\usepackage{booktabs}

\title{ELMER: Evolutionary Language Model that Explores and Refines}

\author{
\begin{minipage}[t]{0.30\textwidth}
    \centering
    \textbf{Matthew Siper}\\[3pt]
    \small
    Nof1, New York University\\
    \texttt{siper.matthew@gmail.com}
\end{minipage}
\hfill
\begin{minipage}[t]{0.30\textwidth}
    \centering
    \textbf{Ahmed Khalifa}\\[3pt]
    \small
    Nof1, University of Malta\\
    \texttt{ahmed.khalifa@um.edu.mt}
\end{minipage}
\hfill
\begin{minipage}[t]{0.30\textwidth}
    \centering
    \textbf{Julian Togelius}\\[3pt]
    \small
    Nof1, New York University\\
    \texttt{julian.togelius@gmail.com}
\end{minipage}
}

\affiliations{}

\begin{document}
\maketitle

\begin{abstract}
Program evolution can measure whether a mutation helped, but it rarely controls how far the mutation moves in behavior space. Syntactic edit size is an unreliable proxy: a small code change can alter nearly every action, while a larger rewrite can preserve the same execution trace. We introduce an Evolutionary Language Model that searches over natural-language policy descriptions and compiles typed programs for execution. A fully fine-tuned Qwen3-8B model learns three task-conditioned operations: conditional semantic mutation, natural language to domain-specific language (GPTL) compilation, and GPTL to natural language translation. The model is fine-tuned with conditional input on the mutation strength (low, medium, high) using Direct Preference Optimization (oDPO). Across 252 fixed-budget evolutionary searches, oDPO improves both behavioral calibration and finite-budget search efficiency. Natural-language attains the highest observed held-out fitness. Our analysis shows that the condition input (mutation strength) systematically changes semantic edit composition and that language mutations preserve more parent fitness at matched small-to-moderate behavioral displacement. These results show that language can serve as a steerable, execution-grounded search representation over executable program space.
\end{abstract}

\section{Introduction}
Program evolution can measure whether a mutation improves fitness, but it
often cannot control how far that mutation moves in behavior space.
Continuous optimizers expose an explicit update scale in that program search
typically relies on syntactic proxies such as token, tree, or subtree edits.
These proxies can be poorly aligned with execution in that a small comparator
change may alter nearly every action a policy takes, while a larger rewrite
may preserve the same action sequence. The missing object is a mutation scale whose requested
magnitude predicts realized behavior.

We ask whether a language model can learn this scale from execution. Our
system separates the representation used for variation from the
representation used for evaluation. Evolution mutates standardized
natural-language (NL) descriptions of trading policies, while a learned
compiler maps each child into typed Genetic Programming Trading Language
(GPTL) code for deterministic execution. Parent and child programs are run
on the same historical market data, and disagreement between their action sequences defines
realized behavioral displacement. A low, medium, or high request targets an
ordered, overlapping behavioral regime.

A single fine-tuned Qwen3-8B large language model (LLM) learns three
task-conditioned operations: NL-to-GPTL compilation, GPTL-to-NL translation,
and conditional NL mutation. We construct mutation supervision from
grammar-valid GPTL degradation chains whose endpoints can be executed and
assigned deterministic behavior distances, then translate those transitions
into language. Behavior-grounded supervised fine-tuning (SFT) learns partial
regime control. Common-parent offset Direct Preference Optimization (oDPO)
sharpens that control by preferring sibling children whose executed
displacement better matches the requested regime. While fitness remains the
outer-loop selection signal, behavioral displacement trains the variation
operator.

We use financial trading as a testbed for stochastic, state-dependent
programmatic control.
Fixed historical trajectories permit deterministic, highly parallel replay
and strict temporal holdouts. Across 252 fixed-budget searches, oDPO
significantly improves mutation-strength calibration and validation-trajectory area
under the curve (AUC) relative to behavior-grounded SFT, native abstract
syntax tree (AST) mutation, and a matched code-output oDPO model. The matched
model holds the backbone, training transitions, preference data,
conditioning, and objective fixed while changing the mutation
representation from language to GPTL code. Natural-language oDPO also discovers
the highest observed held-out policy, while held-out mean and upper-tail
results are treated descriptively. Mechanistic analyses show that requested
strength changes semantic edit composition and that language mutations
preserve more parent fitness at matched small-to-moderate behavioral
displacement.

Our contributions are:
\begin{itemize}
    \item an execution-grounded definition of mutation displacement and a
    conditional variation operator over ordered behavioral regimes;
    \item a multitask language model trained for semantic mutation,
    compilation, and translation using behaviorally calibrated executable
    transitions;
    \item controlled ablations that isolate domain fine-tuning,
    behavior-grounded conditioning, oDPO, and the natural-language
    representation; and
    \item mechanistic analyses of calibration, semantic edit composition,
    syntax--behavior locality, and mutation quality at matched displacement.
\end{itemize}

The rest of the paper is organized as follows. We first review related work
on program evolution, LLM-guided search, preference optimization. We then introduce the behavior-space mutation operator,
describe the multitask training data and oDPO objective, and define the
mutation systems used in evolution. Next, we present the experimental design
and report the calibration, search-efficiency, representation, and
mechanistic results. Finally, we discuss the implications and limitations of
the approach before concluding.

\section{Related Work}

\subsection{Program Evolution, Semantics, and Locality}
Genetic programming (GP) searches executable structures, so representation and variation jointly determine the neighborhood exposed to selection \citep{koza1992genetic,rothlauf2006representations}. Syntactically small edits can produce large output changes, motivating semantic operators \citep{moraglio2012geometric}. MAP-Elites preserves elites across behavioral niches \citep{mouret2015illuminating}. Transformer Semantic GP learns semantically related proposals \citep{anthes2025transformer}. Continuous Program Search learns a behavior-aware continuous space for typed trading programs \citep{siper2026continuous}. We instead learn categorical control directly in the proposal distribution from executed parent--child displacement.

\subsection{LLM-Guided Evolution and Automated Discovery}
Evolution through LLMs established language models as learned variation operators \citep{lehman2022evolution}. Evaluator-guided systems have been employed to evolve functions, model code, heuristics, metaheuristics, and scientific programs \citep{romeraparedes2024funsearch,morris2024llmguidedevolution,liu2024eoh,vanstein2025llamea,novikov2025alphaevolve,adaption2026autoscientist}. ProFiT evolves executable trading programs from historical market feedback \citep{siper2026profit}. These systems evaluate proposals through fitness or execution. We additionally train how far the proposal operator moves in behavior space.

\subsection{Preference Optimization for Behavior-Grounding}
Scalar fitness and validity describe \textit{how well} a program performs, but they fail to capture \textit{how far} a mutation has moved from its parent in behavior space. To resolve this ambiguity, we adapt the reverse-degradation principle from Path of Destruction \citep{siper2022pod}. By reversing synthetic degradation chains, we construct a diverse dataset of directionally improving program transitions across multiple discrete behavioral displacements. We then optimize our mutation operator using a margin-aware preference framework. While standard Direct Preference Optimization (DPO) optimizes for a binary quality gradient \citep{rafailov2023dpo}, oDPO incorporates pair-dependent margins to scale the preference update \citep{amini2024odpo}. By computing these margins directly from the observed trajectory displacement of common-parent sibling pairs, we repurpose preference optimization. Rather than simply ranking output quality, the model learns a steerable, ordinal geometry of the policy mutation space.

\section{A Behavior-Space Mutation Operator}
\subsection{Behavior-Space Step Size}
Let $\mathcal{G}$ be the set of valid executable programs, $\mathcal{R}_{\mathrm{NL}}$ the set of language descriptions, $\mathcal{S}$ the market-state space, and $\mathcal{B}=\{0,1\}^4$ the raw signal space for long entry, short entry, long exit, and short exit. A policy $\pi:\mathcal{S}\rightarrow\mathcal{B}$ maps a market state to a four-signal vector. For a shared evaluation trajectory $\mathbf{s}_{1:N}=(s_1,\ldots,s_N)\in\mathcal{S}^N$, behavioral displacement is equivalent to the raw-signal disagreement:
\begin{equation}
 d_{\mathrm{beh}}(\pi_a,\pi_b)=\frac{1}{N}\sum_{t=1}^{N}
 \mathbb{I}\!\left[\pi_a(s_t)\neq\pi_b(s_t)\right].
 \label{eq:behavior-distance}
\end{equation}
A learned compiler $C:\mathcal{R}_{\mathrm{NL}}\rightarrow\mathcal{G}$ maps child description $x_c$ to a program inducing $\pi_c$. Sampling $x_c\sim p_\theta(\cdot\mid x_p,s)$ and compiling induces $q_\theta(\pi_c\mid\pi_p,s)$ for $s\in\{\mathrm{low},\mathrm{medium},\mathrm{high}\}$. Let $\mu_s$ be the empirical center of regime $s$. An idealized target is
\begin{equation}
 \min_{\theta}\;\mathbb{E}_{\pi_p,s}\,
 \mathbb{E}_{\pi_c\sim q_\theta(\cdot\mid\pi_p,s)}
 \left[\left|d_{\mathrm{beh}}(\pi_p,\pi_c)-\mu_s\right|\right].
 \label{eq:conditional-proposal}
\end{equation}
Because execution is nondifferentiable, common-parent oDPO instead ranks children by proximity to $\mu_s$, subject to
\begin{equation}
 D_{\mathrm{low}}\prec D_{\mathrm{medium}}\prec D_{\mathrm{high}}.
 \label{eq:ordinal-geometry}
\end{equation}
These categories are ordered, overlapping regimes rather than exact numerical radii. Each representation-operator pair induces its own practical reachability distribution.

\begin{figure*}[!t]
    \centering
    \includegraphics[width=0.90\textwidth]{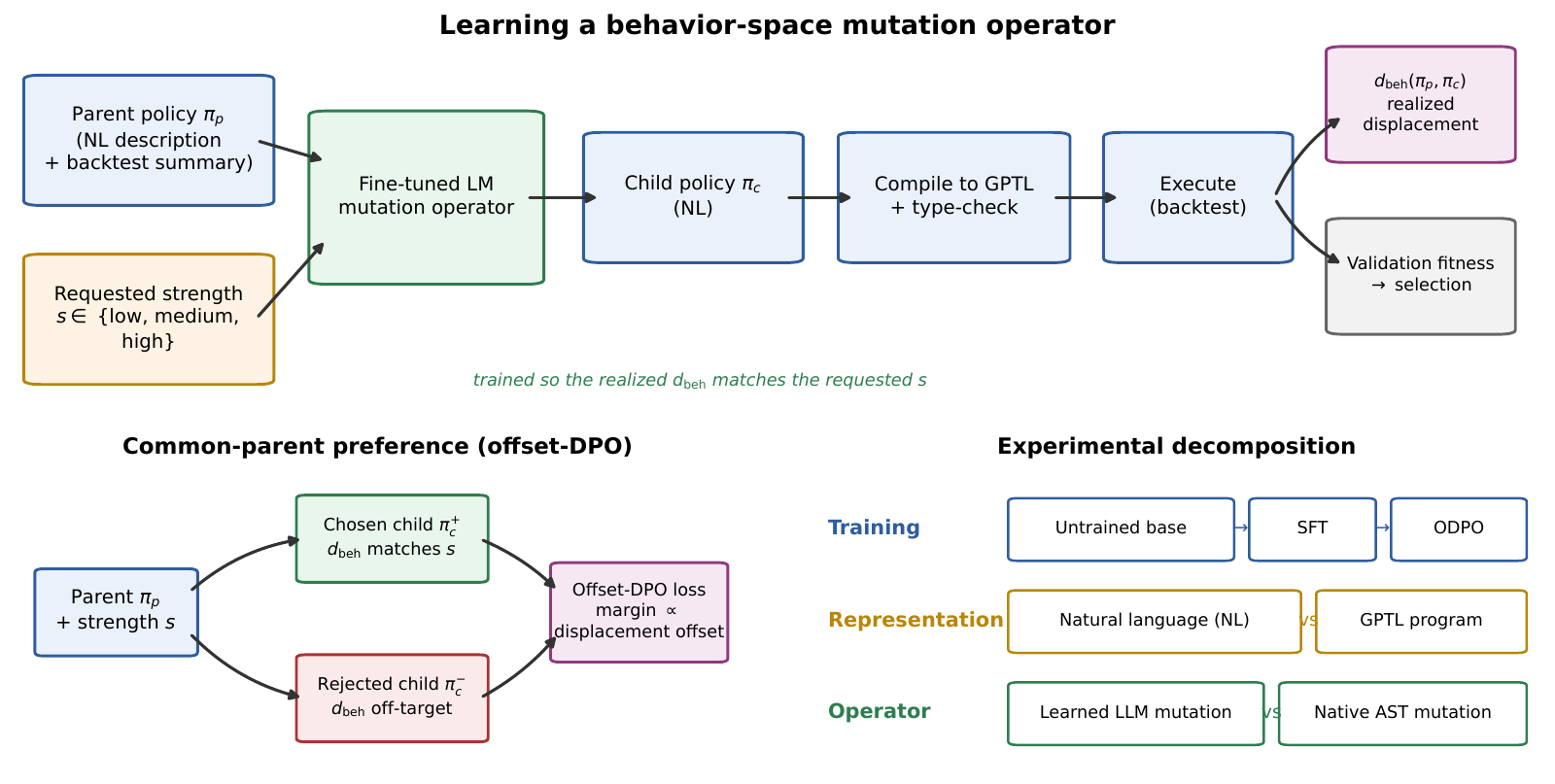}
    \caption{Learning a behavior-space mutation operator. A parent policy and requested strength produce a language mutation that is compiled and executed. Displacement trains the variation operator, while validation fitness drives selection. The lower panels show common-parent oDPO preferences and the experimental decomposition.}
    \label{fig:method}
\end{figure*}

\subsection{Behavior-Grounded Multitask Mutation Model}

We fine-tune a single Qwen3-8B checkpoint \citep{yang2025qwen3} on three
task-conditioned operations. \textit{Compilation} maps a natural-language
policy description to a typed GPTL program that can be checked, executed, and evaluated through backtesting. Program-to-language \textit{translation} maps a GPTL program into the standardized natural-language representation consumed by the semantic mutator. \textit{Mutation} receives a parent description, its backtest feedback, and a requested strength
$s\in\{\mathrm{low},\mathrm{medium},\mathrm{high}\}$, then produces a child description whose executed behavioral displacement should fall within the
corresponding regime. Together with deterministic backtesting and fitness-based selection, these operations form a closed evolutionary loop. Policies are mutated in language, compiled into GPTL code, executed to obtain fitness, and retained or rejected by the outer optimization algorithm. Translation allows a hand-designed GPTL seed, a directly mutated program, or another code-space discovery to enter the language representation used by the mutator. During ordinary natural-language evolution, the child description is inherited directly and its compiled GPTL program is stored as the evaluated artifact, so translation is not repeatedly applied along the lineage.

The same loop provides a controlled way to construct behavior-grounded mutation supervision. Starting from high-fitness GPTL programs, defined as training Sharpe greater than 0.5, we apply grammar-valid AST edits and retain edges that directionally reduce fitness, producing degradation chains. Reversing one-, two-, and four-link segments yields directionally improving parent-child transitions at several mutation scales. Because both endpoints are executable programs, we run them on the same training-only market windows and compute their behavioral displacement using Eq.~\ref{eq:behavior-distance}. GLM 5.2, employed as an annotation model with reasoning off and temperature equal to 0.1, then renders both endpoints in the standardized natural-language format. The resulting examples pair a natural-language semantic edit with a displacement measured deterministically from the original executable programs. The model therefore learns to mutate in language while developing an implicit association between semantic edit patterns and their realized behavioral regimes.

The final supervised corpus contains 45,639 examples: 13,551 conditional mutations, 16,044 compilations, and 16,044 program-to-language translations. These collections are split into 44,727 training and 912 validation examples. The mutation data are derived from 2,811 degradation chains and 44,920 graded reverse transitions, of which 44,917 execute successfully. Strength labels are assigned using equal-frequency behavioral-distance bins with cut points 0.165 and 0.466 as determined by the data distribution. We remove null transitions below 0.005 and transitions near the category
boundaries, then balance the retained examples across strength and asset. Distances are measured on six training-only windows of 6,000 hourly bars. Mutation prompts provide the available domain primitives, the parent policy (both its GPTL and NL representations), its fitness, and seven backtest statistics. Conditioned prompts add the requested mutation strength.

\subsection{Common-Parent Preferences and oDPO}
For each eligible parent, we select one representative child per strength category. Given request $s$, we designate as preferred $y^+$ the child nearest $\mu_s$ and as rejected $y^-$ a child outside the requested category. We filter out pairs whose behavioral separation falls below $0.15$ times the training interquartile range (IQR) to ensure a distinct preference gap. Each retained pair receives
\begin{equation}
 m=\min\!\left(1,\frac{|d^+-d^-|}{Q_{0.95}(|d^+-d^-|)}\right).
 \label{eq:margin}
\end{equation}
Here $Q_{0.95}$, the 95th percentile of training separations, robustly scales typical pairs. Clipping at one prevents extreme outliers, from dominating the oDPO offset and gradient. The procedure yields 14,735 training and 775 validation pairs from 6,369 parents.

Let $r_\theta(y\mid x)=\log\pi_\theta(y\mid x)-\log\pi_{\mathrm{ref}}(y\mid x)$, where $\pi_{\mathrm{ref}}$ is the frozen behavior-grounded SFT model. We optimize
\begin{equation}
\mathcal{L}_{\mathrm{oDPO}}
=-\mathbb{E}\!\left[\log\sigma\!\left(\beta\Delta r_\theta-\lambda m\right)\right],
\label{eq:odpo}
\end{equation}
with $\Delta r_\theta=r_\theta(y^+\mid x)-r_\theta(y^-\mid x)$, $\beta=0.3$, and $\lambda=0.2$. SFT uses one epoch at $2\times10^{-5}$; oDPO uses one epoch at $2\times10^{-6}$. Checkpoints must exceed 95\% on compilation validity, translation signal preservation, and output format, and pass a held-out monotonicity gate with Spearman $\rho\geq0.4$.

\subsection{Mutation Systems and Evolution}
The ablation ladder separates four effects. Stock Qwen3-8B measures
zero-shot mutation; no-label SFT isolates domain and mutation-task
fine-tuning; prompt-only labels test whether strength words alone induce control; behavior-grounded SFT tests supervised conditioning on measured displacement; and oDPO isolates the additional effect of common-parent preference optimization after conditional SFT. Two code controls isolate representation and operator effects. The native baseline compares learned language mutation with a hand-designed weighted AST operator spanning parameter, indicator, subtree, comparator, logical, clause, crossover, and signal-leg edits. This same operator system was used to generate the degradation chains that ultimately produced the multi-task training dataset. The matched code-output model holds the backbone, training transitions, preference pairs, strength labels, and oDPO objective fixed while changing the output substrate from natural language to GPTL. It therefore provides the cleanest test of natural language as a search representation.

\section{Experimental Design}
Experiments use hourly continuous futures from January 2008 through October 2025: the E-mini S\&P 500 contract (ES; 106,685 bars), Silver (SI; 107,085), and U.S. Treasury Bond (US; 105,124). The backtester starts with \$10,000, applies 0.005\% commission and 0.01\% slippage, and uses unit long, short, or flat positions. Each asset has three non-overlapping temporal folds, split chronologically into 70\% training, 15\% validation, and 15\% test, with a ten-day embargo. During search, the mutation model receives training-fold feedback; validation scores drive evolutionary selection, and the best-validation policy is evaluated once on the corresponding held-out test fold after the 1,000-evaluation budget ends. The learned-operator matrix contains six operators, three algorithms, three assets, and four seeds (216 searches). A separate 36-search AST matrix supplies the native baseline, and the 36 NL oDPO runs are shared across analyses. We report validation Sharpe, held-out test Sharpe, top-quartile mean test Sharpe, maximum test Sharpe, validity, and area under the best-so-far validation curve on the common 1,000-evaluation grid (AUC).

Lineage analysis measures normalized AST token distance, Eq.~\ref{eq:behavior-distance}, edit taxonomy, entropy, and fitness change in shared behavioral-distance bins. A mutation is neutral when $d_{\mathrm{beh}}=0$ and catastrophic when an execution-valid child's validation Sharpe falls by more than one; failures are tracked separately through validity. Runs are paired by algorithm, asset, and seed, yielding 36 matched observations per contrast. Six confirmatory contrasts use two-sided Wilcoxon signed-rank tests with Holm correction, 10,000-resample paired-bootstrap intervals, and rank-biserial effects. Held-out test fitness is the endpoint for the three capability and conditioning contrasts; AUC is the endpoint for the three oDPO and representation contrasts. Representation-level mean, top-quartile, and maximum test fitness are descriptive. Calibration uses run-level Spearman correlations followed by paired Wilcoxon tests with Holm correction; pooled low--high Cliff's $\delta$ is descriptive.

\section{Results}
\subsection{A Requested Strength Becomes a Behavioral Control}
\begin{figure*}[!t]
    \centering
    \includegraphics[width=0.94\textwidth]{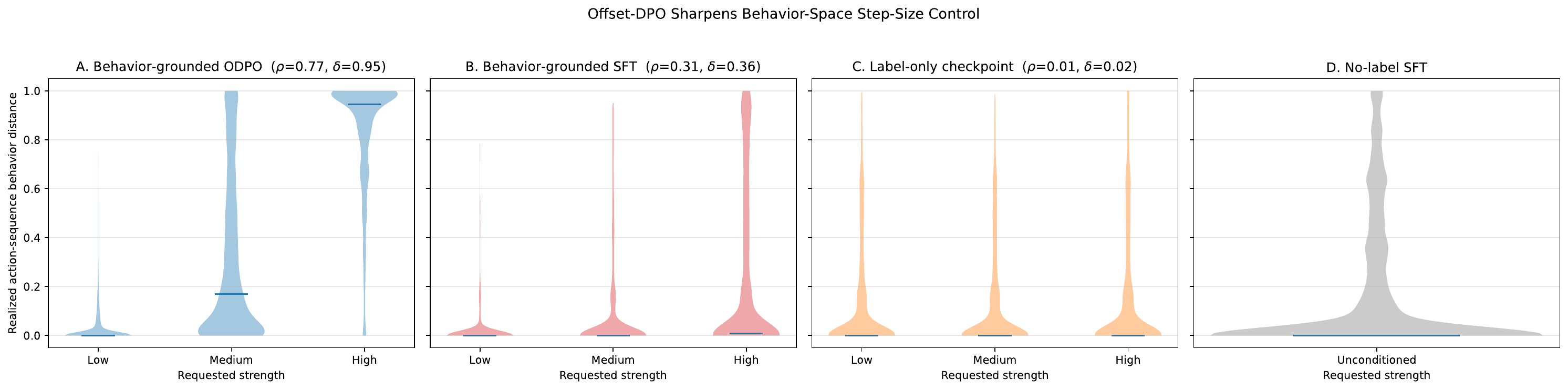}
    \caption{Requested strength versus realized action-sequence displacement. Behavior-grounded SFT learns partial ordering, oDPO sharpens it, and prompt-only labels produce no measurable response. Bars mark medians.}
    \label{fig:calibration}
\end{figure*}

Figure~\ref{fig:calibration} shows the pooled ordinal response. In the run-level analysis used for inference, mean Spearman correlation between requested strength and realized displacement is $0.824$ for oDPO, $0.310$ for behavior-grounded SFT, and $0.018$ for prompt-only conditioning. Every one of the 36 paired runs favors oDPO in both comparisons ($r_{\mathrm{rb}}=1.00$, $p_{\mathrm{Holm}}<0.0001$). Descriptive pooled low--high Cliff's deltas are $0.98$, $0.36$, and $0.02$. oDPO therefore turns partial supervised ordering into consistent regime control; the broad medium distribution and high-strength saturation still preclude exact numerical calibration.

\subsection{Where the Mutation Capability Comes From}
\begin{figure*}[!t]
    \centering
    \includegraphics[width=0.90\textwidth]{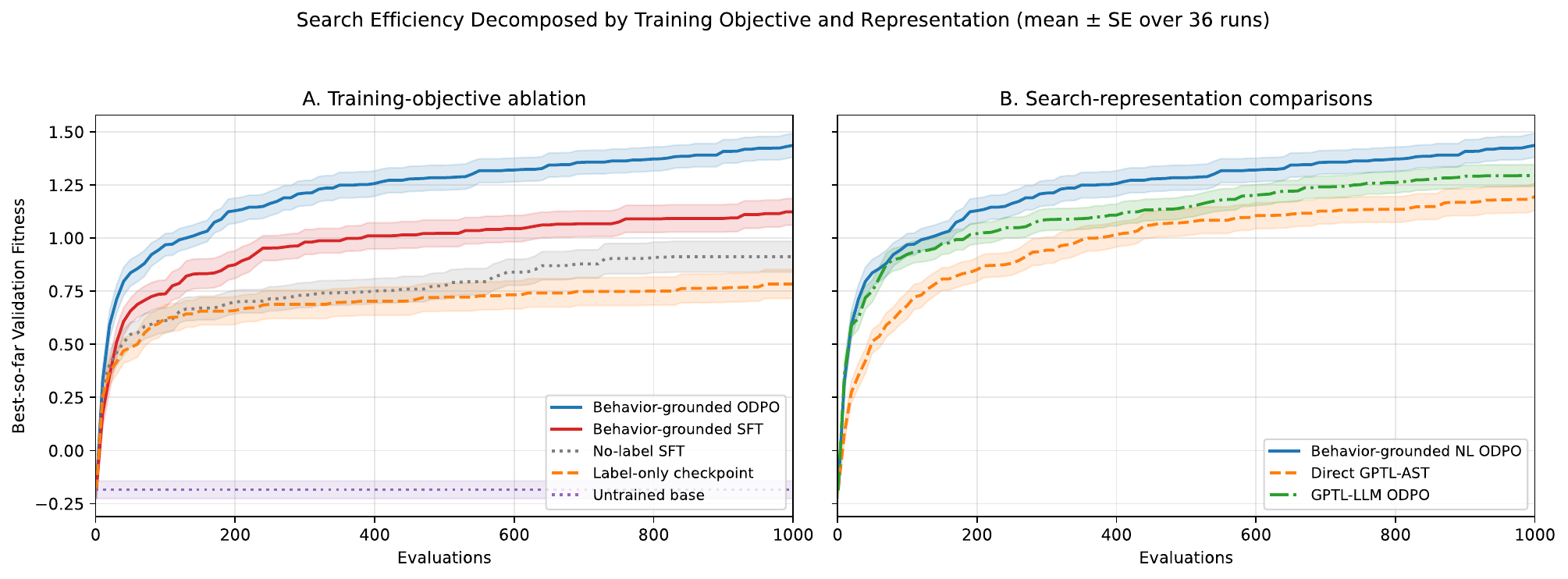}
    \caption{Best-so-far validation fitness. Panel A decomposes training. Panel B compares NL with native AST mutation and a matched code-output oDPO model. Curves show mean $\pm$ standard error (SE) over 36 runs.}
    \label{fig:search-efficiency}
\end{figure*}

Table~\ref{tab:main-results} and Figure~\ref{fig:search-efficiency}A decompose the operator. The untrained model yields negative validation, test, and AUC. No-label SFT becomes functional. Prompt-only labels lower every average metric relative to no-label SFT, showing that grounding gives the conditioning interface its operational meaning.

\begin{table*}[!t]
\centering
\small
\setlength{\tabcolsep}{3.8pt}
\renewcommand{\arraystretch}{0.92}
\begin{tabular}{@{}lrrrrr@{}}
\toprule
\multicolumn{6}{l}{\textbf{A. Descriptive aggregates (36 searches per system)}} \\
Mutation system & \shortstack{Mean\\Validation} & \shortstack{Mean\\Test} & \shortstack{Top-Quartile\\Test} & \shortstack{Maximum\\Test} & \shortstack{Mean\\AUC} \\
\midrule
\multicolumn{6}{l}{\emph{Natural-language training ablation}} \\
Untrained base & -0.1844 & -0.2163 & 0.2745 & 0.4027 & -109.0 \\
No-label SFT & 0.9122 & 0.1583 & 0.8021 & 1.1715 & 491.1 \\
Label-only checkpoint & 0.7836 & -0.0575 & 0.5059 & 1.3383 & 436.9 \\
Behavior-grounded SFT & 1.1232 & \textbf{0.4805} & \textbf{1.5386} & 1.9841 & 527.9 \\
Behavior-grounded NL oDPO & \textbf{1.4364} & 0.4478 & 1.5063 & \textbf{2.3445} & \textbf{751.2} \\
\midrule
\multicolumn{6}{l}{\emph{Representation controls}} \\
Direct code (AST) & 1.1949 & 0.3526 & 0.9959 & 1.2598 & 569.9 \\
Code-output oDPO & 1.2938 & 0.3024 & 1.1967 & 1.6319 & 693.0 \\
\midrule
\multicolumn{6}{l}{\textbf{B. Confirmatory paired contrasts (36 matched runs)}} \\
Contrast & Endpoint & $\Delta$ & 95\% CI & $r_{\mathrm{rb}}$ & $p_{\mathrm{Holm}}$ \\
\midrule
No-label SFT $-$ Base & Test & $+0.375$ & $[0.189,\ 0.559]$ & $0.64$ & \textbf{0.0024} \\
Label-only $-$ No-label & Test & $-0.216$ & $[-0.417,\ 0.003]$ & $-0.51$ & \textbf{0.0251} \\
BG-SFT $-$ Label-only & Test & $+0.538$ & $[0.220,\ 0.869]$ & $0.56$ & \textbf{0.0111} \\
NL oDPO $-$ BG-SFT & AUC & $+223.3$ & $[134.8,\ 320.0]$ & $0.82$ & \textbf{0.0001} \\
NL oDPO $-$ Direct AST & AUC & $+181.3$ & $[85.0,\ 295.1]$ & $0.58$ & \textbf{0.0078} \\
NL oDPO $-$ Code oDPO & AUC & $+58.2$ & $[2.3,\ 123.9]$ & $0.39$ & \textbf{0.0425} \\
\bottomrule
\end{tabular}
\caption{Training, representation, and confirmatory results. Panel A reports descriptive aggregates. Panel B reports paired mean differences and bootstrap intervals; $p_{\mathrm{Holm}}$ comes from two-sided Wilcoxon signed-rank tests with Holm correction across the six contrasts. Test denotes held-out fitness and AUC denotes evaluation-index search-curve area. Representation-level held-out summaries and maxima remain descriptive.}
\label{tab:main-results}
\end{table*}

Panel B shows that domain SFT creates a functional mutator, prompt-only labels significantly degrade performance, and behavior-grounded SFT restores a useful conditional interface. All three oDPO AUC contrasts remain significant after Holm correction, establishing its primary gain in finite-budget search efficiency. Panel A provides descriptive held-out context: behavior-grounded SFT has the highest mean and top-quartile test fitness, while NL oDPO attains the highest observed test fitness ($2.3445$). The label-only bootstrap interval targets the paired mean, whereas its Wilcoxon test targets a paired rank shift, so the interval can narrowly cross zero while the corrected rank test remains significant.

\begin{figure*}[!t]
    \centering
    \includegraphics[width=0.82\textwidth]{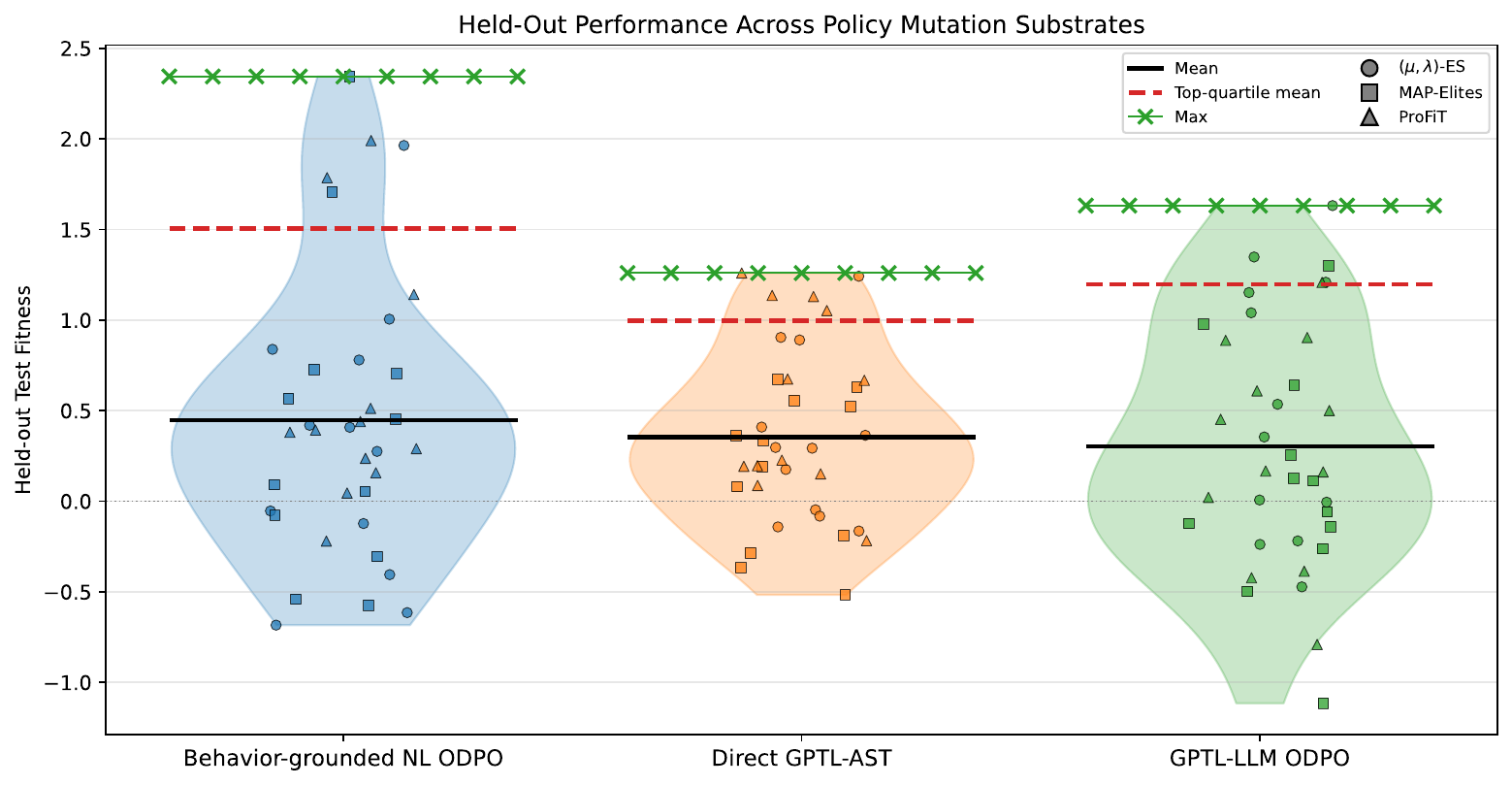}
    \caption{Held-out test fitness across representation--operator systems. Points are searches; marker shape denotes the outer algorithm. Lines show mean, top-quartile mean, and maximum.}
    \label{fig:test-distribution}
\end{figure*}

\begin{figure*}[!t]
    \centering
    \includegraphics[width=0.88\textwidth]{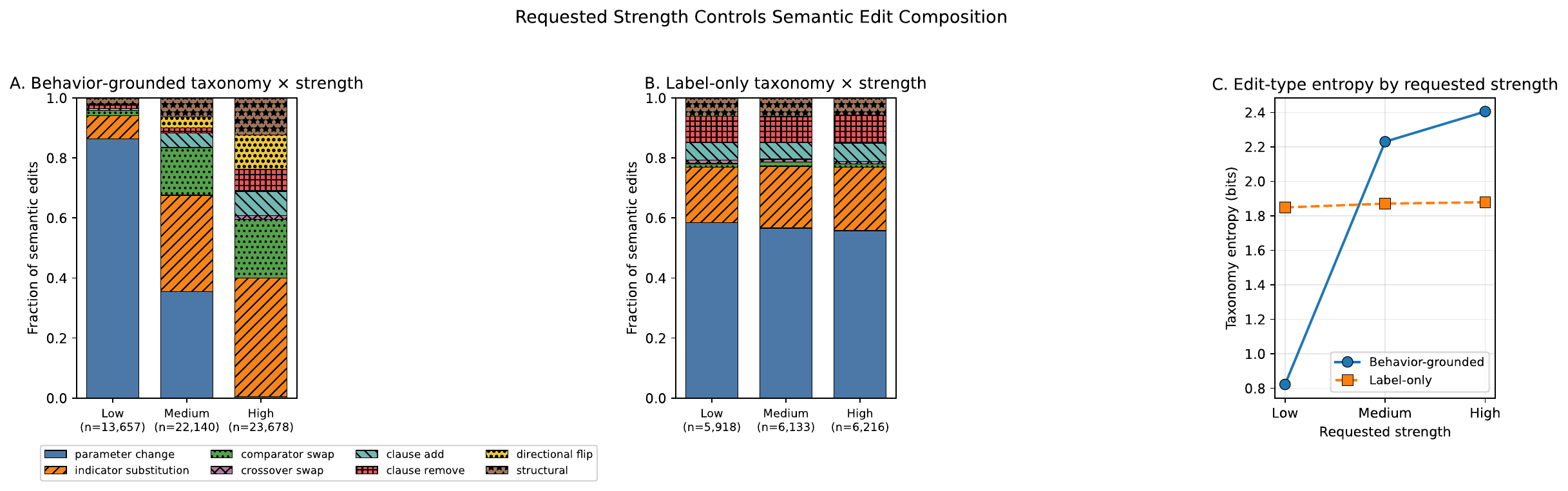}
    \caption{Realized edit composition by requested strength. The behavior-grounded operator changes edit types and entropy, while the prompt-only checkpoint remains nearly invariant.}
    \label{fig:taxonomy}
\end{figure*}

\begin{figure*}[!t]
    \centering
    \includegraphics[width=0.84\textwidth]{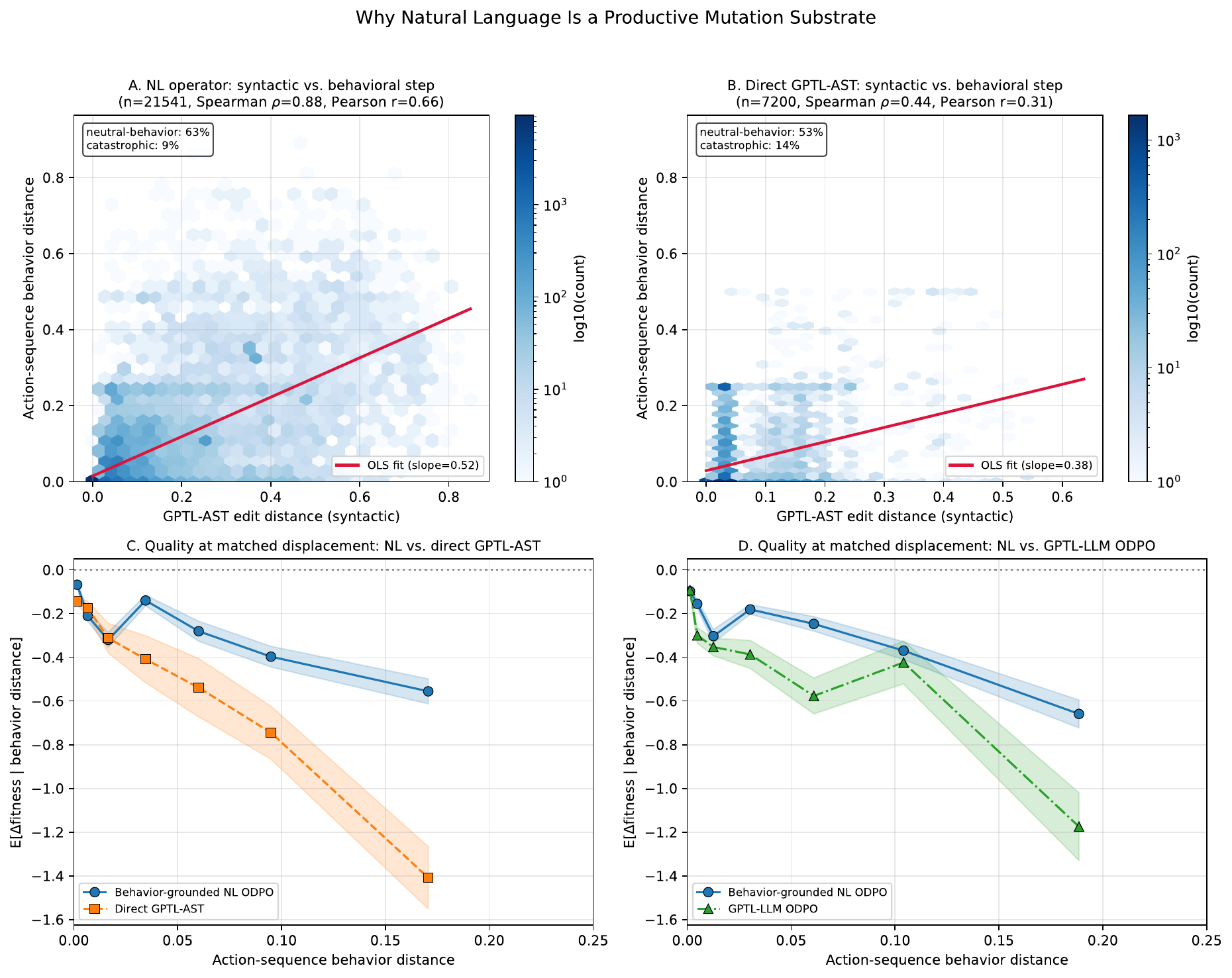}
    \caption{Why language is a productive mutation substrate. Panels A--B relate AST distance to executed behavior for NL and native code mutations. Panels C--D compare fitness change at matched displacement for $d_{\mathrm{beh}}\leq0.25$.}
    \label{fig:substrate-mechanisms}
\end{figure*}

\subsection{Language Improves Finite-Budget Search Efficiency}
Figure~\ref{fig:search-efficiency}B and Panel B of Table~\ref{tab:main-results} compare NL oDPO with native AST mutation and the matched code-output control. Both AUC contrasts remain significant after Holm correction. Because the matched model holds the backbone, transitions, preferences, labels, and objective fixed, its contrast isolates the representation used to express variation. The code-output effect is modest and near the corrected threshold, but the two comparisons support a finite-budget search-efficiency advantage for language. Figure~\ref{fig:test-distribution} provides descriptive held-out context. NL oDPO has higher mean test fitness than native AST ($0.4478$ versus $0.3526$) and matched code output ($0.4478$ versus $0.3024$), as well as the highest top-quartile mean ($1.5063$) and observed maximum ($2.3445$). These values indicate a stronger observed upper tail but are not confirmatory representation endpoints. All systems include negative outcomes, and the outer-loop pattern is heterogeneous: MAP-Elites and ProFiT favor NL on mean test, whereas $(\mu,\lambda)$-ES favors matched code output. Archive memory may help preserve NL's structured exploration, but the experiment does not isolate this interaction.

\subsection{Strength Selects Different Mutation Regimes}

Figure~\ref{fig:taxonomy} shows that parameter changes constitute 86.4\% of low-strength edits, 35.5\% at medium, and 0.46\% at high. High strength reallocates probability toward indicator substitutions (39.5\%), comparator swaps (19.7\%), structural rewrites (12.4\%), directional flips (11.4\%), and clause operations. Entropy rises from $0.82$ to $2.23$ to $2.40$ bits, while prompt-only remains near $1.86$ bits. Requested strength therefore selects distinct semantic mutation regimes rather than merely changing surface wording.

\subsection{Why Language Produces More Useful Moves}
Compiled NL mutations have a stronger syntax--behavior relationship ($\rho=0.88$, $r=0.66$) than native AST mutation ($\rho=0.44$, $r=0.31$) and a lower catastrophic rate (9\% versus 14\%). Catastrophic denotes an execution-valid child losing more than one validation Sharpe; failures are counted separately. Both mappings remain noisy. Within the common-support range $d_{\mathrm{beh}}\leq0.25$, NL preserves more parent fitness than native AST mutation and generally more than matched code output, supporting more coherent moves at comparable executed distance.

\section{Discussion}

Low, medium, and high mutation-strengths acquire operational meaning through executed
parent--child displacement. Prompt-only labels leave behavior nearly
unchanged, while behavior-grounded SFT establishes partial ordering, and oDPO makes
that ordering consistent across matched runs. The stages play complementary
roles. SFT creates a functional semantic mutator and yields the strongest
descriptive mean and top-quartile held-out performance, while oDPO primarily
improves calibration and finite-budget search efficiency. The strongest
representation-level evidence is validation-trajectory AUC, where NL oDPO
significantly exceeds native AST mutation and matched code-output oDPO, while its
higher held-out mean, upper tail, and held-out maximum remain descriptive. Because the
matched control fixes the backbone, training transitions, preference pairs,
labels, and objective, this contrast most directly tests the representation
used for variation. Mechanistic results indicate that language
produced more behaviorally ordered and less damaging small-to-moderate moves.
Variation across outer loops and uniform strength sampling motivate adaptive
mutation scheduling as a future research focus.

\section{Conclusion}

Program evolution has long been able to reward a mutation after the fact but has been far less able to control how far that mutation moves before
evaluation. We address this gap by separating semantic variation from
deterministic execution: a multitask language model edits policy intent in natural language, compiles each proposal into typed GPTL, and learns mutation strength from the action trajectories produced by execution. Behavior-grounded SFT creates the conditional mutation capability, while common-parent oDPO
turns low, medium, and high requests into reliable ordinal behavioral regimes
and significantly improves finite-budget search efficiency over
behavior-grounded SFT, native AST mutation, and a matched code-output model.
The resulting operator also discovers the highest observed held-out policy,
while mechanistic analyses show that requested strength reshapes semantic edit
composition and preserves more parent fitness at matched small-to-moderate
behavioral displacement. These results move program evolution beyond blind
syntactic perturbation toward deliberate, execution-grounded movement through
program space with models that learn not only what to change, but how far to move.

\clearpage
\bibliography{references_aaai2027}
\end{document}